\documentclass[]{bytedance_seed}
\usepackage{tabularx}
\usepackage[toc,page,header]{appendix}

\usepackage{minitoc}
\usepackage{cleveref} 
\usepackage{subcaption}
\usepackage{booktabs}
\usepackage{graphicx}
\usepackage{CJKutf8}
\usepackage{pgfplots}
\usepackage{pgfplotstable}
\usepackage{xcolor}
\usepackage{CJKutf8}
\usetikzlibrary{patterns}
\usepackage{float}
\usepackage{caption}
\newcommand{\method}{StructVRM\xspace}
\usepackage{listings}
\usepackage{xcolor}
\usepackage{tcolorbox}
\tcbuselibrary{listingsutf8}

\lstdefinelanguage{json}{
    basicstyle=\ttfamily\small,
    numbers=none,
    stepnumber=1,
    numbersep=8pt,
    showstringspaces=false,
    breaklines=true,
    frame=single,
    backgroundcolor=\color{gray!5},
    literate=
     *{0}{{{\color{black}0}}}{1}
      {1}{{{\color{black}1}}}{1}
      {2}{{{\color{black}2}}}{1}
      {3}{{{\color{black}3}}}{1}
      {4}{{{\color{black}4}}}{1}
      {5}{{{\color{black}5}}}{1}
      {6}{{{\color{black}6}}}{1}
      {7}{{{\color{black}7}}}{1}
      {8}{{{\color{black}8}}}{1}
      {9}{{{\color{black}9}}}{1}
      {:}{{{\color{black}:}}}{1}
      {,}{{{\color{black},}}}{1}
      {"}{{{\color{black}"}}}{1}
}
\newtcolorbox{promptbox}[1][]{
  colbacktitle=black!60,
  coltitle=white,
  fontupper=\footnotesize,
  colback=gray!5,
  boxsep=5pt,
  left=5pt,
  right=5pt,
  top=5pt,
  bottom=5pt,
  boxrule=1pt,
  title={#1},
  width=0.95\textwidth, % 控制宽度
  enhanced,
  breakable,            % 支持自动分页和换行
}

\usepackage{natbib}
\usepackage{latexsym}

\usepackage{url}
\usepackage{amssymb}
\usepackage[utf8]{inputenc}
\usepackage{microtype}
\usepackage{booktabs}
\usepackage{pifont} 
\usepackage{multirow}
\usepackage{makecell}
\usepackage{paralist}
\usepackage{xspace}
\usepackage{color}
\usepackage{xcolor}
\usepackage{colortbl}
\usepackage{adjustbox}
\usepackage{hyperref} 
\usepackage[edges]{forest}
\usepackage{tikz} 
\usepackage{caption}
\usepackage{amsfonts}

\hypersetup{
    colorlinks,
    linkcolor={blue!80!black},
    citecolor={blue!80!black},
}
\tikzset{
    root/.style =             {align=center, text width=1cm, rounded corners=3pt, line width=0.3mm, fill=gray!10, draw=gray!80, font=\small},
    demographic/.style =         {align=center, text width=1.8cm, rounded corners=3pt, line width=0.3mm, fill=blue!10, draw=blue!80, font=\footnotesize},
    demographic_work/.style =    {align=center, text width=10cm, rounded corners=3pt, line width=0.3mm, fill=blue!10, draw=blue!0, font=\footnotesize},
    character/.style =         {align=center, text width=1.8cm, rounded corners=3pt, line width=0.3mm, fill=red!10, draw=red!80, font=\footnotesize},
    character_work/.style =    {align=center, text width=10cm, rounded corners=3pt, line width=0.3mm, fill=red!10, draw=red!0, font=\footnotesize},
    personalization/.style =           {align=center, text width=1.8cm, rounded corners=3pt, line width=0.3mm, fill=cyan!10, draw=cyan!80, font=\footnotesize},
    personalization_work/.style =      {align=center, text width=10cm, rounded corners=3pt, line width=0.3mm, fill=cyan!10, draw=cyan!0, font=\footnotesize},
    risk/.style =         {align=center, text width=1.8cm, rounded corners=3pt, line width=0.3mm, fill=orange!10, draw=orange!80, font=\footnotesize},
    risk_work/.style =    {align=center, text width=10cm, rounded corners=3pt, line width=0.3mm, fill=orange!10, draw=orange!0, font=\footnotesize},
}

\usepackage{CJK}
\title{\method: Aligning Multimodal Reasoning with Structured and  Verifiable Reward Models}

\affiliation{ByteDance Seed China}

\contribution{Full author list in \hyperref[sec:contributions]{Contributions}}
\abstract{

Existing Vision-Language Models often struggle with complex, multi-question reasoning tasks where partial correctness is crucial for effective learning. Traditional reward mechanisms, which provide a single binary score for an entire response, are too coarse to guide models through intricate problems with multiple sub-parts.
To address this, we introduce StructVRM, a method that aligns multimodal reasoning with Structured and Verifiable Reward Models. At its core is a model-based verifier trained to provide fine-grained, sub-question-level feedback, assessing semantic and mathematical equivalence rather than relying on rigid string matching. This allows for nuanced, partial credit scoring in previously intractable problem formats. Extensive experiments demonstrate the effectiveness of StructVRM. Our trained model, Seed-StructVRM, achieves state-of-the-art performance on six out of twelve public multimodal benchmarks and our newly curated, high-difficulty STEM-Bench. The success of StructVRM validates that training with structured, verifiable rewards is a highly effective approach for advancing the capabilities of multimodal models in complex, real-world reasoning domains.

\date{July 13, 2025}
}

\begin{document}

\maketitle

%不需要目录就注释掉 注意目录不要和第一页放在一块 要有\newpage
%\newpage
%\tableofcontents
%\newpage

\section{Introduction}

The recent proliferation of multimodal reasoning models has marked a significant milestone in artificial intelligence, demonstrating remarkable capabilities in understanding and integrating information across vision and text. Existing models\citep{guo2025seed1,o1,gemini2.5} have achieved impressive performance on a variety of general-purpose tasks, such as visual question answering, image captioning, and open-world dialogue. 
% However, their proficiency often diminishes when confronted with complex reasoning multi-question that require deep, logical deduction, particularly in specialized domains like science, technology, engineering, and mathematics (STEM). 
However, their proficiency often diminishes when confronted with complex, multi-question reasoning tasks that require deep logical deduction, particularly in specialized domains like science, technology, engineering, and mathematics (STEM).
These tasks demand more than surface-level pattern recognition; they necessitate a robust grasp of procedural logic, symbolic manipulation, and the ability to interpret intricate diagrams and formulas.

\begin{figure}[t]
    \centering
    \small
    \includegraphics[width=\linewidth]{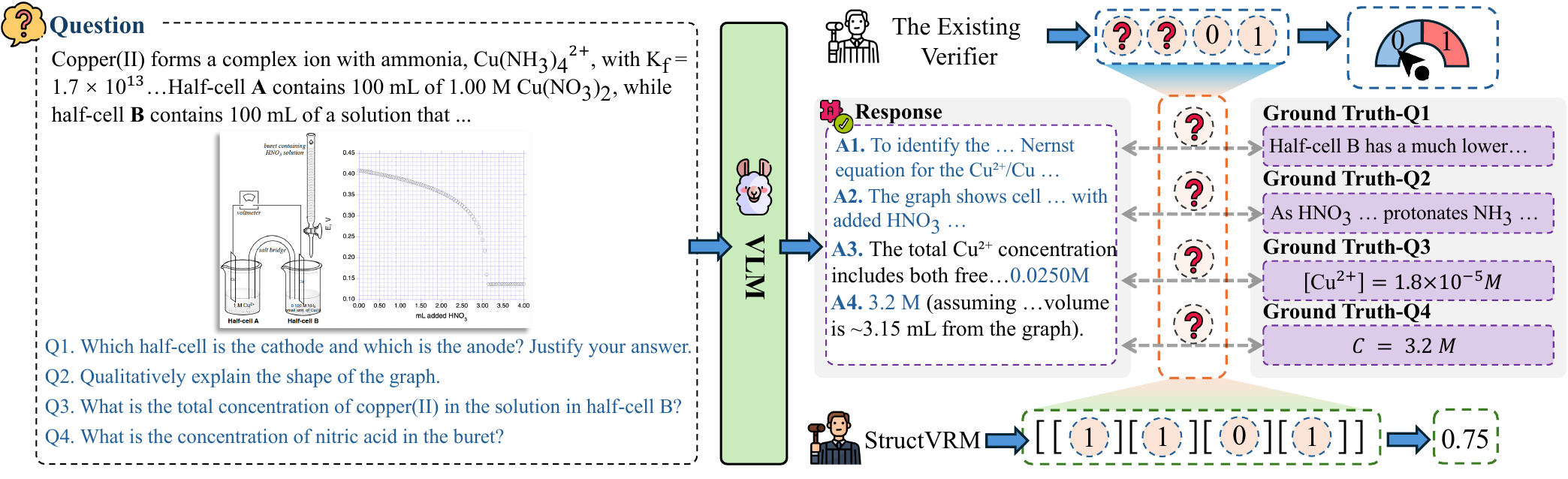}
    % \vspace{-8pt}
    % \caption{MLLMs can only verify final answers but fail to assess intermediate reasoning steps.}
    \caption{A four-part electrochemistry problem with open-ended sub-questions (Q1–Q2) and simple numeric sub-questions (Q3–Q4). The existing verifier assigns only independent binary scores to the numeric items and cannot validate open-ended responses; StructVRM evaluates all sub-questions and outputs a structured, fine-grained score vector.}
    \label{fig:intro-verify-gap}
    \vspace{-10pt}
\end{figure}

% A fundamental bottleneck in advancing multimodal reasoning is the nature of the feedback signals used during training, especially in reinforcement learning (RL). 
A fundamental bottleneck in advancing multimodal reasoning is the nature of the feedback signals used during training, especially in reinforcement learning with verifiable rewards (RLVR). 
% Conventional RL frameworks rely on coarse, scalar rewards—typically a binary signal indicating whether a final answer is entirely correct or incorrect. 
Conventional RLVR methods, as depicted in Figure~\ref{fig:intro-verify-gap}, rely on coarse, scalar rewards—typically a binary signal indicating whether a final answer is entirely correct or incorrect.
This "all-or-nothing" approach is profoundly inefficient for structured problems with multiple sub-questions. A model that correctly solves three out of four sub-problems receives the same zero-reward penalty as a model that fails completely, providing no gradient for partial progress. This sparse and uninformative feedback makes it difficult for the model to learn from its mistakes and incrementally build complex reasoning chains, often leading to slow convergence or policy collapse.
% Moreover, existing rule based verifiers can only validate simple single item numeric answers and they fail to handle real world responses that are structurally complex and varied in format. 

% \vspace{-20pt}

Moreover, existing rule based verifiers can only validate simple single item numeric answers and they fail to handle real world questions that are structurally complex and varied in format. 
% In real-world scenarios, data often present high complexity, placing greater demands on verification mechanisms and calling for more fine-grained approaches. 
% In real-world scenarios, data are often highly complex.
Our analysis of actual exam data shows that approximately 62.88\% of the questions are verifiable, such as multiple-choice and single-answer fill-in-the-blank questions. The remaining 37.12\% consist of questions that are hard-to-verify and non-verifiable, including multi-part fill-in-the-blank, complex reasoning, and open-ended subjective questions. Although reference answers are provided, the open and diverse nature of responses makes simple binary verification inadequate. 
% Existing verification and training methods struggle to fully utilize such data, limiting model generalization on complex tasks.
% In real-world settings, data often exhibit a high degree of complexity, placing greater demands on verification mechanisms and highlighting the need for more sophisticated and fine-grained verification frameworks. For example, our analysis of actual examination data indicates that approximately 62.88\% of the questions are of a verifiable nature, such as multiple-choice and single-answer fill-in-the-blank questions. However, around 37.12\% of the questions fall into the category of difficult-to-verify or unverifiable types, including multi-part fill-in-the-blank items, complex reasoning problems, and open-ended subjective questions. Although reference answers are typically provided for these questions, their open-ended and diverse formats make them challenging to verify through simple binary feedback mechanisms. Current mainstream verification and training approaches struggle to make effective use of such data, limiting the generalization capability of models when handling complex tasks and hindering progress toward improved performance in more realistic and context-rich scenarios.

To overcome these limitations, we introduce StructVRM, a method designed to align multimodal reasoning with structured and  verifiable reward. Instead of relying on a simplistic binary judgment, StructVRM employs a sophisticated, model-based verifier that evaluates model-generated answers at a granular level. This verifier is trained to parse structured outputs and provide a score vector, assessing the correctness of each sub-question independently. By learning to recognize semantic and mathematical equivalence, our verifier moves beyond brittle, exact-match evaluation to provide a more robust and meaningful feedback signal.

The Seed-StructVRM training pipeline is a two-stage process designed to build and refine complex reasoning abilities. First, we perform Supervised Fine-Tuning (SFT) on a high-quality dataset of over 50,000 multimodal problems, each paired with a detailed Chain-of-Thought (CoT) reasoning trace. This phase instills a strong foundation for generating structured, step-by-step answers. Second, we apply RL using Proximal Policy Optimization (PPO), where the policy is guided by the rich, structured rewards from our verifier. This verifier-guided RL stage allows the model to refine its reasoning pathways, rewarding partial progress and steering it toward complete and accurate solutions.

To rigorously evaluate our approach, we created a comprehensive data ecosystem, including a new, challenging benchmark named STEM-Bench, which features high-difficulty questions from math, physics, chemistry, and biology that require deep reasoning. Our contributions are threefold:

\begin{itemize}
    % \item We propose the StructVRM method, which pioneers the use of a model-based verifier to generate structured, fine-grained rewards for training complex reasoning models.
    \item We propose StructVRM, a method that leverages a model-based verifier to assign structured, fine-grained rewards and extends RLVR to complex multi-question and open-ended tasks, enhancing real-world generalization and data utilization efficiency.
    % \item We construct a large-scale, high-quality dataset for multimodal reasoning and introduce STEM-Bench, a novel benchmark designed to push the boundaries of scientific problem-solving in VLMs.
    % \item We further extend the RLVR method to more complex multi-question scenarios, enabling reinforcement learning to generalize effectively to data that better reflects real-world applications and improving data utilization efficiency.
    \item We construct a large-scale, high-quality dataset for multimodal reasoning and introduce STEM-Bench, a novel benchmark designed to push the boundaries of scientific problem-solving in VLMs.
    % \item We demonstrate through extensive experiments that StructVRM achieves state-of-the-art performance across 13 public benchmarks and significantly outperforms leading proprietary models like o3 and Gemini 2.5 Pro on STEM-Bench, underscoring the effectiveness of our verifier-aligned reward strategy.
    % \item We demonstrate through extensive experiments that StructVRM achieves state of the art performance across thirteen public benchmarks and outperforms state of the art models on STEM Bench, underscoring the effectiveness of our verifier aligned reward strategy.
    % \item We demonstrate through extensive experiments that StructVRM achieves state-of-the-art  performance on six of twelve public benchmarks, including +5.3 on VLM2 Bench and +3.2 on RealworldQA, and sets a new SOTA of 79.23 on STEM Bench, underscoring our verifier-aligned reward strategy.
    \item We demonstrate through extensive experiments that our trained model, Seed-StructVRM, achieves state-of-the-art performance on six out of twelve public multimodal benchmarks and STEM-Bench, underscoring the effectiveness of our verifier-aligned reward strategy.

\end{itemize}

Our work represents a significant step toward building more capable and reliable multimodal reasoning, proving that structured, verifiable feedback is a key ingredient for unlocking the next level of intelligent behavior.

\section{Related Work}

% O1. R1. QwQ. Grok. Gemini-thinking. Claude-3.7 thinking

% \subsection{Reasoning Models}

% \subsection{Reinforcement Learning}

% Test-time scaling~\cite{qwq,grok,gemini-thinking,claude3.7} such as OpenAI’s o1~\cite{o1} and DeepSeek’s R1~\cite{r1} have catalyzed a profound paradigm shift in LLMs~\cite{gpt3,gpt4}. By enabling extended CoT reasoning~\cite{cot} and eliciting sophisticated reasoning capabilities, these methods empower LLMs to excel in complex mathematical and coding tasks, including those from competitions like the AIME and Codeforces.
% At the core of this transformation is large-scale reinforcement learning, which facilitates the emergence of complex reasoning behaviors—such as self-verification and iterative refinement. 
% However, the critical methodologies and algorithms underpinning scalable RL training have largely remained obscure, often omitted from the technical documentation of existing reasoning models~\cite{o1,r1,gpt3,gpt4,cot}.
% In this paper, we introduce an SOTA-level model \method and introduce the details to achieve the performance from three aspects: Data, RL algorithm, and RL infrastructure.

\subsection{Multimodal STEM Benchmarks}

Recent advances in multimodal reasoning have prompted the development of diverse benchmarks spanning mathematics, physics, chemistry, and biology. Existing datasets primarily fall into three categories based on domain coverage and modality structure. The first group comprises math-focused multimodal benchmarks that combine diagrammatic and textual input to evaluate models' visual mathematical reasoning. MathVista~\cite{lu2023mathvista}, MathVerse~\cite{zhang2024mathverse}, MathVision~\cite{wang2025solidgeo}, We-Math~\cite{qiao2024we}, Polymath~\cite{gupta2024polymath}, and MathScape~\cite{zhou2024mathscape} offer varying degrees of question formats including multiple-choice, open-ended, and diagram-based queries. While these datasets cover a range of math subdomains and include thousands of problems, they often lack fine-grained structural annotations or difficulty calibration across question types, limiting their use for structured reward modeling or hierarchical verification. The second group includes science-level or cross-disciplinary benchmarks such as R-Bench~\cite{li2024r}, OlympiadBench~\cite{he2024olympiadbench}, SciBench~\cite{wang2024scibench}, SceMQA~\cite{liang2024scemqa}, and VNHSGE~\cite{dao2023vnhsge}, which span mathematics, physics, chemistry, and biology. These datasets generally offer large-scale and multilingual coverage, but their multimodal components often consist of loosely integrated visual and textual elements, and rarely support fine-grained control over task type or reasoning complexity. Finally, vertical efforts in specific subjects, such as ChemVLM~\cite{li2025chemvlm}, ChemRxivQuest~\cite{amiri2025chemrxivquest}, and evaluations in biology reasoning~\cite{nguyen2023evaluating}, focus on domain-specific knowledge extraction or molecular-level interpretation. Although these datasets address specific challenges, existing benchmarks typically lack multi-question scenarios due to difficulties in evaluation. Across all categories, existing benchmarks underrepresent tasks that demand tightly coupled multimodal interpretation, structured CoT-style annotations, or difficulty-controlled reasoning setups. 
In contrast, our benchmark bridges these gaps by specifically collecting challenging, real-world multimodal reasoning data from complex multi-question scenarios.

% Although these datasets contribute targeted challenges, they tend to be smaller in scale and lack generalizable question transformation protocols. Across all categories, existing benchmarks underrepresent tasks that demand tightly coupled multimodal interpretation, structured CoT-style annotations, or difficulty-controlled reasoning setups. In contrast, our benchmark construction addresses these gaps by generating high-difficulty, multimodal reasoning problems with structured CoT rationales and automated question-type conversions (e.g., choice-to-blank or judgment), enhancing reasoning complexity and reducing answer space ambiguity.

\subsection{Verifier-Based Reward Design}

Verifier-based reward mechanisms play a central role in aligning large language and vision-language models with structured reasoning tasks. Existing approaches can be broadly categorized into three major types: answer-based verifiable rewards, structure- and heuristic-driven rewards, and model-informed semantic verifiers. The first category centers on tasks where gold-standard answers exist and can be directly compared against model predictions. This includes binary correctness rewards such as those in R1-V~\cite{zhang2025r1}, math verification tools used in Reason-RFT~\cite{tan2025reason}, multiple-choice scoring functions from VLM-R1~\cite{shen2025vlm}, and IoU or F1-based metrics applied in Visual-RFT~\cite{liu2025visual} and Perception-R1~\cite{yu2025perception}. These methods are computationally efficient and widely adopted but are inherently limited to tasks with fully deterministic outputs. 
To address tasks involving intermediate reasoning or generation structure, several works propose heuristic-based rewards that evaluate output format, length, or redundancy. VisualThinker-R1-Zero~\cite{zhou2025r1} enforces structured output via \texttt{<think>} and \texttt{<answer>} tags, while Kimi 1.5~\cite{team2025kimi}, Kimi-VL~\cite{team2025kimi}, and DAPO~\cite{yu2025dapo} apply length-sensitive bonuses or penalties to encourage informative yet concise outputs. 
Additional strategies such as n-gram repetition penalties~\cite{yeo2025demystifying}, regular expression-based step counting~\cite{faceopen}, and language constraint filters~\cite{ma2025rethinking} further refine structural fidelity. However, these handcrafted metrics often fail to capture reasoning validity or semantic faithfulness. To bridge this gap, recent efforts explore model-informed verifiers that leverage pretrained language models or reflection mechanisms. VAPO~\cite{yue2025vapo} utilize large language models to assign semantic scores or integrate policy-gradient and NLL loss to reward rare correct generations; VL-Rethinker~\cite{wang2025vl} introduces rethinking prompts and rewards corrected second-pass answers. 
% While diverse in formulation, these methods typically treat evaluation and supervision as separate stages, and few attempt to integrate evaluation directly into the training signal. 
In contrast, our work introduces the StructVRM method, featuring a Verifiable Reward Model that unifies answer-level correctness, sub-question-level semantic consistency, and structural format validation into a single reward system. This model supports rule-based verification for deterministic tasks and model-based equivalence assessment for hard-to-verify and non-verifiable  problems, enabling interpretable, fine-grained supervision across diverse multimodal reasoning formats.

\subsection{Reward Modeling for Multimodal Reasoning}

Recent progress in aligning large language models with complex reasoning tasks has led to the development of diverse reward modeling strategies. These approaches can be broadly categorized into three classes based on their supervision structure: reinforcement learning with verifiers (RLVR), process-level reward modeling (PRM), and generative reward modeling (GRM). RLVR methods such as DeepSeek-R1~\cite{guo2025deepseek} and Kimi-1.5~\cite{team2025kimi} employ rule-based or structured verifiers to assign scalar or format-sensitive rewards, while RLPR~\cite{yu2025rlpr} removes external evaluators by introducing a likelihood-based proxy reward and GRPO-LEAD~\cite{zhang2025grpo} introduces difficulty-aware weighting to improve gradient quality across diverse problems. In contrast, PRM-based approaches like Math-Shepherd~\cite{wang2023math}, rStar-Math~\cite{guan2025rstar}, and R-PRM~\cite{she2025r} emphasize step-level feedback by supervising intermediate reasoning steps, often using Monte Carlo tree search or automatic scoring. Extensions to multimodal domains include MM-PRM~\cite{du2025mm}, which incorporates symbolic and visual signals, and ReasonFlux-PRM~\cite{zou2025reasonflux}, which models the coherence of entire reasoning trajectories. A third direction is GRM, which reframes reward modeling as a generative task. JudgeLM~\cite{zhu2023judgelm} fine-tunes large models to act as preference judges, while RM-R1~\cite{chen2025rm}, GRAM~\cite{wang2025gram}, and ReasonGRM~\cite{chen2025reasongrm} combine ranking objectives with reasoning-aware generation to enhance reward interpretability and generalization. While these methods have shown strong performance in various settings, they often suffer from limitations in symbolic reasoning, visual-textual alignment, or modular integration. In contrast, our proposed StructVRM method combines rule-based scoring for verifiable tasks with model-based structured evaluation for complex and open-ended problems, enabling fine-grained and scalable supervision across diverse multimodal reasoning settings.
% 第29行服务于我们自己故事的引出，这里需要谭铖师兄看下，这里需不需要改下

\section{Dataset Construction}
\label{sec:Dataset Construction}

\begin{figure}[!h]
\centering
\includegraphics[width=\textwidth]{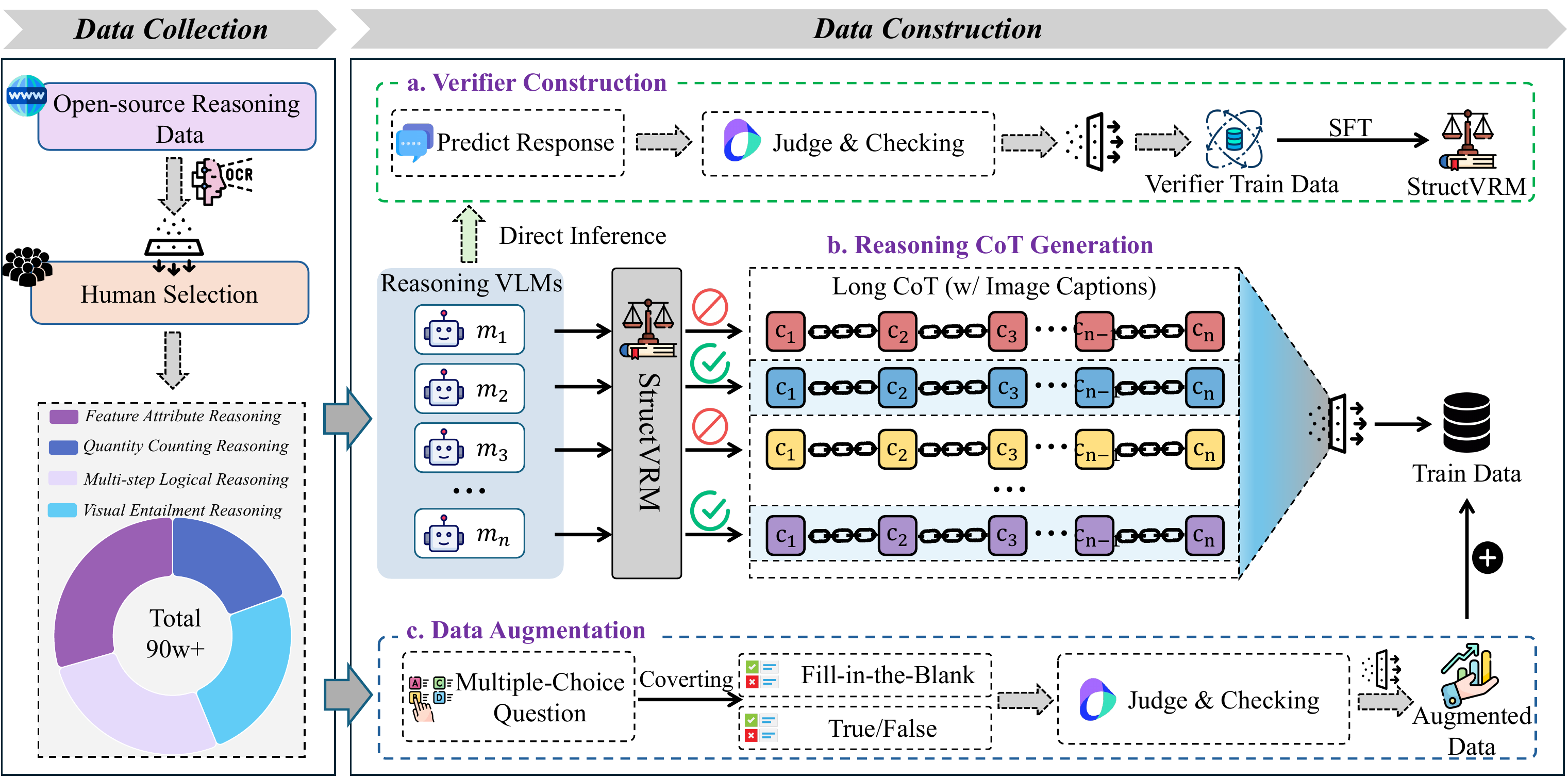}
\caption{Overview of the dataset construction pipeline.}
\label{fig:dataset-pipeline}
\end{figure}

Developing general-purpose multimodal reasoning systems for real-world tasks requires high-quality, structurally diverse, and instruction-compatible data. To address this, we construct a comprehensive data pipeline aimed at collecting, structuring, and enriching multimodal reasoning problems involving deep visual-textual understanding and multi-step logic. This pipeline supports both supervised and reinforcement learning stages, with structured verification and reward routing as integral components. As shown in Figure~\ref{fig:dataset-pipeline}, our dataset construction process includes two main phases: (1) large-scale collection and organization of multimodal problems, and (2) high-quality reasoning data construction including verifier building, reasoning chain generation, and format diversification.

\subsection{Overview and Scope}

We begin by collecting large-scale, publicly available multimodal questions that involve diagrams, formulas, procedural reasoning, symbolic structures, and complex multi-question scenarios data. These problems reflect real-world task demands and are characterized by high structural complexity, visual-textual dependencies, and diverse reasoning formats.
% We begin by collecting large-scale, publicly available multimodal questions that involve diagrams, formulas, procedural reasoning, and symbolic structures. These problems reflect real-world task demands and are characterized by high structural complexity, visual-textual dependencies, and diverse reasoning formats.
From this corpus, we curate over 50,000 high-quality problems for supervised fine-tuning (SFT), including multiple-choice, fill-in-the-blank, and open-ended formats. All samples are filtered through OCR-based parsing, multi-stage cleaning, and manual review to ensure clear semantics, correct answers, and well-aligned visual content.

To enhance reward modeling and long-form robustness, we further construct a dedicated set of over 100,000 challenging examples designed for reinforcement learning. These instances are selected using a verifier-guided difficulty estimator, prioritizing problems with long reasoning chains, symbolic ambiguity, and visually grounded inference.

In parallel, we curate a specialized verifier training set of over 200,000 structured examples, each annotated with sub-question correctness. 
% This dataset supports learning a fine-grained reward model capable of decomposed evaluation across reasoning steps.
This dataset supports learning a fine-grained reward model capable of sub-question decomposition evaluation.

To increase reasoning diversity and enable structured feedback learning, we divide the overall construction pipeline into three components:
\begin{itemize}
    \item \textbf{Verifier Construction}: Train a neural verifier to assess multi-question correctness using structured and scalable evaluation strategies.
    \item \textbf{CoT Reasoning Generation}: Leverage multiple internal vision-language models (VLMs) to generate image-aware, multi-hop chain-of-thought responses for each question.
    \item \textbf{Data Augmentation}: Convert a question from one format to another (e.g., split a single multiple-choice item into several true/false statements) to boost model generalization and curb shortcut bias.

    % Convert problems across formats (e.g., MCQ $\rightarrow$ fill-in, MCQ $\rightarrow$ true/false) to improve generalization and reduce shortcut bias.
\end{itemize}

\subsection{High-Quality Dataset Construction}
The StructVRM data pipeline integrates three core modules: (a) Verifier Construction, (b) Reasoning CoT Generation, and (c) Task-Type Data Augmentation. These components support different stages of training: CoTs serve as supervised learning targets, the verifier model drives reward scoring in reinforcement learning, and data augmentation promotes structural diversity.

\subsubsection{Verifier Construction}

We implement model-based scoring in StructVRM to provide structured correctness signals for complex answers. This reward model is trained on over 200,000 annotated examples, each scored at the sub-question level using interpretable JSON outputs.

The annotation process begins with multiple VLMs generating diverse answers for each problem. These responses are evaluated by a internal large language model (LLM) using a strict scoring rubric that enforces segment-wise correctness, binary judgments, and standardized format. The full grading prompt is illustrated in Figure~\ref{fig:verifier-prompt}. This results in a training set that maps (question, student answer, reference answer) to structured score vectors, capturing both partial correctness and semantic/mathematical equivalence. To further enhance data quality, we introduce a rigorous filtering and verification pipeline, which includes format validity checks, content completeness reviews, semantic consistency verification, and the removal of low-quality samples. These steps aim to minimize scoring bias and hallucination noise during training, ensuring that the supervision signals remain clear and reliable.

To assess the quality of the verifier model, we constructed a held-out evaluation set annotated by human experts. On this benchmark, the verifier achieves 96.83\% agreement with expert judgments, demonstrating its strong alignment with human evaluation and reliability in reward modeling.
% The model-based scorer is not used to filter supervised samples, but instead provides fine-grained reward signals for reinforcement learning.

\begin{figure}[H]
\centering
\resizebox{\textwidth}{!}{
\begin{CJK}{UTF8}{gbsn}
\begin{promptbox}[Verifier Prompt]
\ttfamily
Now your role is that of a strict grading teacher. Your task is to review and score the student’s answers with reference to the standard answers. Throughout the grading process, you need to be thoroughly familiar with the following key points:\\
% - Grading should be based solely on the student’s final answers; you do not need to assess the correctness of the intermediate solution steps.\\
% - First extract the final answers from the student’s solutions and display them in your analysis, then evaluate whether they are correct.\\
- The questions you are grading consist of multiple sub-questions; therefore, you need to evaluate each sub-question individually.\\
% - Grade based on your analysis, and when stating your scoring rationale, describe it in segments according to your logical analysis. The summary of your scoring rationale should be placed at the end, using a format such as: “In summary, the student’s answers deserve x points” (where x represents the student’s specific score).\\
- Provide the grades according to your analysis and present them in a code block in JSON format.\\
Please strictly adhere to the output format requirements. Your output format is: \\
【Scoring Basis】:\\
【Total Score】:X points\\
【JSON】：\\
\{\\
    "score": [[Score]]\\
\}\\
% Where Score is always given as a list, e.g. [1,0]; each sub-question is represented by a list, and if that sub-question has multiple parts, use 0 or 1 within that list accordingly. \\
% For example: A question has 3 sub-questions; the first has 3 parts, the second has 1 part, the third has 2 parts. Then the Score format is: ``score":[[1,0,1],[1],[1,0]]. Note! For multiple-choice questions, only the final selected options versus the standard answer are considered: each question is scored with a single point—1 if all options are correct, 0 if any are missing or incorrect.\\
% Below is the grading rubric：\\
% 【Score Levels】:Evaluate the final answer against the standard answer, with two levels: 1 point and 0 points (the minimum is 0; if a 0-point situation would require further deduction, still give 0 points).\\
【Level Details】:\\
1 point：\\
- The student’s final answer matches the standard answer; award 1 point.\\
% - For multiple-choice questions, if the final selected options are all correct, award 1 point.\\
% - If a question has multiple sub-questions, score each sub-question separately: correct = 1, incorrect = 0, and output a score list for each question, e.g. [1,0,1].\\
% - The student’s answer and the standard answer are mathematically equivalent (e.g., student writes 1+1/2x and standard is 1+0.5x; since 1/2 = 0.5, award 1 point).\\
- The student’s answer and the standard answer are mathematically equivalent.\\
% - The student’s answer and the standard answer are semantically equivalent (e.g., student: ``Esterification is reversible and slow, requiring concentrated sulfuric acid as a catalyst"; standard: ``Esterification is a reversible reaction with a slow rate, requiring concentrated sulfuric acid as a catalyst"; since both express the same meaning, award 1 point).\\
- The student’s answer and the standard answer are semantically equivalent.\\
0 points：\\
% - For multiple-choice questions, any missing or incorrect selection yields 0 points.\\
- The student’s final answer is neither semantically nor mathematically consistent with the standard answer; give 0 points.\\
【Example】\\
Question: \\
\{question\}

Standard Answer: \\
\{standard\_answer\}

Student Answer: \\
\{student\_answer\}

\end{promptbox}
\end{CJK}
}
% \caption{Prompt used for structured scoring in Seed-STEM-Verifier annotation and training.}
\caption{Prompt used for structured scoring in StructVRM annotation and training.}
\label{fig:verifier-prompt}
\end{figure}

\subsubsection{Reasoning CoT Generation}

% To support supervised fine-tuning, we generate long-form, image-aware reasoning traces. For a curated set of 51,254 problems, four vision-language models (Thinking 3.9 RL, 3.8 RL, 3.8 SFT, and 1.0 RL) are each prompted multiple times to produce diverse chains-of-thought.
To support supervised fine-tuning, we generate long-form, image-aware reasoning traces. For a curated set of 51,254 problems, multiple internal models are each prompted multiple times to produce diverse chains-of-thought.

Outputs are filtered using heuristics and our constructed StructVRM, retaining completions that include boxed final answers, accurate visual grounding, and logically coherent steps. These reasoning traces cover various types of multimodal inference, including attribute comparison, quantity counting, multi-hop deduction, and entailment.

\subsubsection{Data Augmentation}

To improve generalization and reduce overfitting to fixed formats, we perform structured format rewriting over selected multiple-choice questions. These conversions allow the model to engage with questions in alternative formats, fostering more flexible reasoning and finer-grained learning signals.

Our augmentation pipeline includes two strategies:

\begin{itemize}
    \item \textbf{Choice $\rightarrow$ Fill-in-the-Blank}: We remove all options from the original question and rewrite the stem as a cloze-style query. The correct answer is used to fill in the response slot. This form eliminates distractors and emphasizes precise generation.
    \item \textbf{Choice $\rightarrow$ True/False}: For proposition-type questions, each option is decomposed into an independent true/false statement. This creates sub-questions that can be judged and trained individually, enhancing the model’s capacity for localized verification.
\end{itemize}

% Both formats are automatically rewritten via prompt-based annotation. Answer keys are propagated using deterministic rules.
Both formats are automatically rewritten via prompt-based annotation, with answer keys propagated using deterministic rules.

\begin{figure}[!h]
\centering
\includegraphics[width=\linewidth]{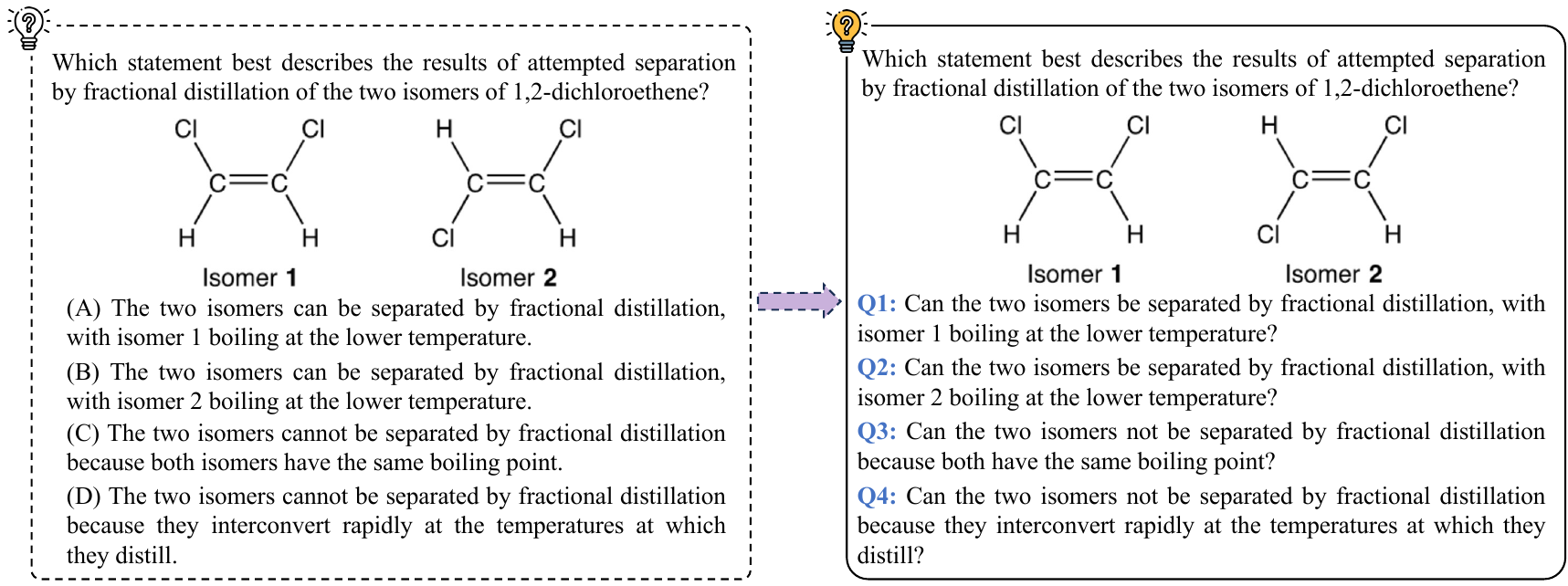}
\caption{Data augmentation – decomposition example}
\label{fig:rewrite-example-decompose}
\vspace{-4px}
\end{figure}

\begin{figure}[!h]
\centering
\includegraphics[width=\linewidth]{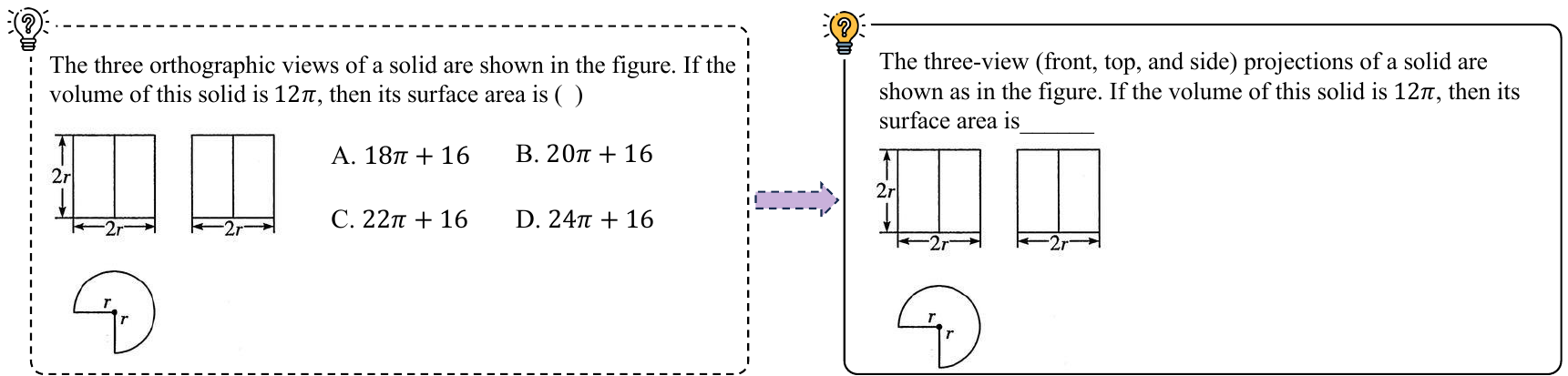}
\caption{Data augmentation – rewriting example}
\label{fig:rewrite-example-rewrite}
\end{figure}

To illustrate how these augmentations work in practice, Figure~\ref{fig:rewrite-example-decompose} and Figure~\ref{fig:rewrite-example-rewrite} present two examples extracted from real exam questions. 
The first showcases a single-choice question with four propositions transformed into a set of independent true/false judgments—each addressing one of the original choices. 
The second demonstrates the conversion from a multiple-choice format to a fill-in-the-blank style, where all distractors are removed and the question is rephrased with a blank answer slot. 
% The first showcases a single-choice question with four propositions transformed into a set of independent true/false judgments—each addressing one of the original choices. 
These augmentations help prevent reward hacking during reinforcement learning by breaking coarse-grained correctness into modular and verifiable units, ensuring the reward model responds to genuine reasoning rather than pattern exploitation.

% \begin{figure}[!h]
% \centering
% \includegraphics[width=\linewidth]{images/chaiti.pdf}
% \caption{Data augmentation – decomposition example}
% \label{fig:rewrite-example-decompose}
% \end{figure}

\section{Model-Based Scoring in StructVRM}
\label{sec:verifier}

We introduce a model-based verifier as part of our StructVRM, responsible for assessing complex, partially-verifiable, or open-ended answers through structured, learnable evaluation rather than rule-based matching. Instead of merely checking final answers against references, this approach formulates verification itself as a trainable subtask, enabling structured, context-aware, and semantically grounded assessment.

Existing scalar verifiers typically return a single score indicating whether a predicted answer matches a reference, which is often too coarse for structured problems where partial correctness matters. For example, in a multi-subquestion exam problem, a model may answer one sub-question correctly and another incorrectly. A scalar verifier would treat this as a total failure, masking the nuance of partial understanding. Worse, when such binary scores are used as rewards in reinforcement learning, they inject noise and degrade learning efficiency.

To address this limitation, we design a model-based scoring module within StructVRM, which provides fine-grained, instruction-tuned verification for structured answers with multiple subcomponents. Instead of mapping predictions to a scalar score, it outputs a vector of sub-question-level scores:
\begin{equation}
\mathbf{s} = f_\theta(\hat{y}, y) = [s_1, s_2, \dots, s_j]
% \mathbf{s} = f_\theta(\hat{y}, y) = [s_1, s_2, \dots, s_k], \quad s_j \in \{0,1\}
\label{eq:score_vector}
\end{equation}
where $\hat{y}$ is the model prediction, $y$ is the reference answer, and each $s_j$ is a one-dimensional list representing the correctness scores of the $j$-th sub-question(s), with each element being either 0 or 1, as determined by semantic or numerical equivalence. Each $s_j$ may correspond to a single sub-question or multiple blanks within a sub-question. The verifier is trained using triplets $\langle \hat{y}, y, \mathbf{s} \rangle$ distilled from a high-quality prompt-based annotation process (see Figure~\ref{fig:verifier-prompt}), where strong LLMs judge correctness based on the final answer only, explicitly ignoring intermediate reasoning to avoid hallucination and procedural noise.

% where $\hat{y}$ is the model prediction, $y$ is the reference answer, and each $s_j$ denotes the correctness of the $j$-th sub-question, determined by semantic or numerical equivalence. The verifier is trained using triplets $\langle \hat{y}, y, \mathbf{s} \rangle$ distilled from a high-quality prompt-based annotation process (see Figure~\ref{fig:prompt}), where strong LLMs judge correctness based on the final answer only, explicitly ignoring intermediate reasoning to avoid hallucination and procedural noise.

% The model adopts a regression-style learning objective over the predicted score vectors:
% \begin{equation}
% \mathcal{L}_{\text{verifier}} = \frac{1}{N} \sum_{i=1}^N \left\| f_\theta(\hat{y}_i, y_i) - \mathbf{s}_i \right\|^2
% \label{eq:verifier_loss}
% \end{equation}
% This encourages the model to learn nuanced correspondences between predicted and reference answers at the sub-question level, and to generalize across formats and domains. Crucially, all inputs to the model are flattened natural language sequences, with no reliance on symbolic program traces or rule-based comparisons.

In downstream reinforcement learning, the model-based verifier provides a structured reward signal by computing:
\begin{equation}
R_{StructVRM} = \frac{1}{k} \sum_{j=1}^k \text{mean}(s_j)
\label{eq:reward}
\end{equation}
which captures the proportion of sub-questions correctly solved. This scalar reward guides the policy toward partial correctness rather than all-or-nothing behavior, enabling more stable and interpretable optimization in complex problem settings. Moreover, since the evaluation is applied to the full answer context, it retains global coherence while offering local feedback—a key property for tasks that involve dependency across sub-questions.

The design of the model-based verifier reflects a shift from rigid symbolic evaluation to neural verification, enabling scalable and nuanced supervision for diverse reasoning tasks in StructVRM.
\section{Reward Modeling}
\label{sec:reward-modeling}

Effective reward modeling is central to reinforcement learning, especially for complex reasoning tasks where feedback must reflect both correctness and partial understanding. In our StructVRM method, we adopt a task-aware reward strategy by categorizing problems into three types based on their verifiability: (1) \textit{verifiable}, where answers can be precisely judged via rule-based matching; (2) \textit{hard-to-verify}, where answers contain structured outputs with partial correctness; and (3) \textit{non-verifiable}, where semantic and mathematical equivalence must be inferred. Verifiable problems are handled with a deterministic rule-based function, while the latter two types are evaluated using our model-based verifier (see Section~\ref{sec:verifier}).

\subsection{Reward Modeling for Verifiable Problems}

For multiple-choice and other highly structured problems where answers follow predictable formats and can be automatically extracted, we apply a rule-based reward function. Specifically, if the predicted answer $\hat{y}$ exactly matches the reference $y$, the reward is $1$; otherwise, $0$:
\begin{equation}
R_{\text{rule}}(\hat{y}, y) =
\begin{cases}
1 & \text{if } \hat{y} = y \\
0 & \text{otherwise}
\end{cases}
\end{equation}
This rule-based method is low-cost, deterministic, and supports rapid convergence in reinforcement learning due to its noiseless feedback.

\subsection{Reward Modeling for Hard-to-Verify and Non-Verifiable Problems}

% For more complex reasoning tasks—including multi-blank fill-ins and open-ended short answer questions—rule-based approaches are inadequate due to expression diversity and the need for partial credit. To address this, we employ the model-based verifier described in Section~\ref{sec:verifier}, which evaluates the predicted answer through fine-grained, structure-aware judgment at the sub-question level.
% \mathbf{s} = f_\theta(\hat{y}, y) = [s_1, s_2, \dots, s_k], \quad s_j \in \{0,1\}
% The verifier outputs a structured score vector:
% \begin{equation}
% \mathbf{s} = f_\theta(\hat{y}, y) = [s_1, s_2, \dots, s_k]
% \mathbf{s} = f_\theta(\hat{y}, y) = [s_1, s_2, \dots, s_k], \quad s_j \in \{0,1\}
% \end{equation}
% where each $s_j$ represents the correctness of the $j$-th sub-question. The final reward used for RL is then computed as the normalized mean of the score vector:

\begin{figure}[!h]
\centering
\includegraphics[width=\linewidth]{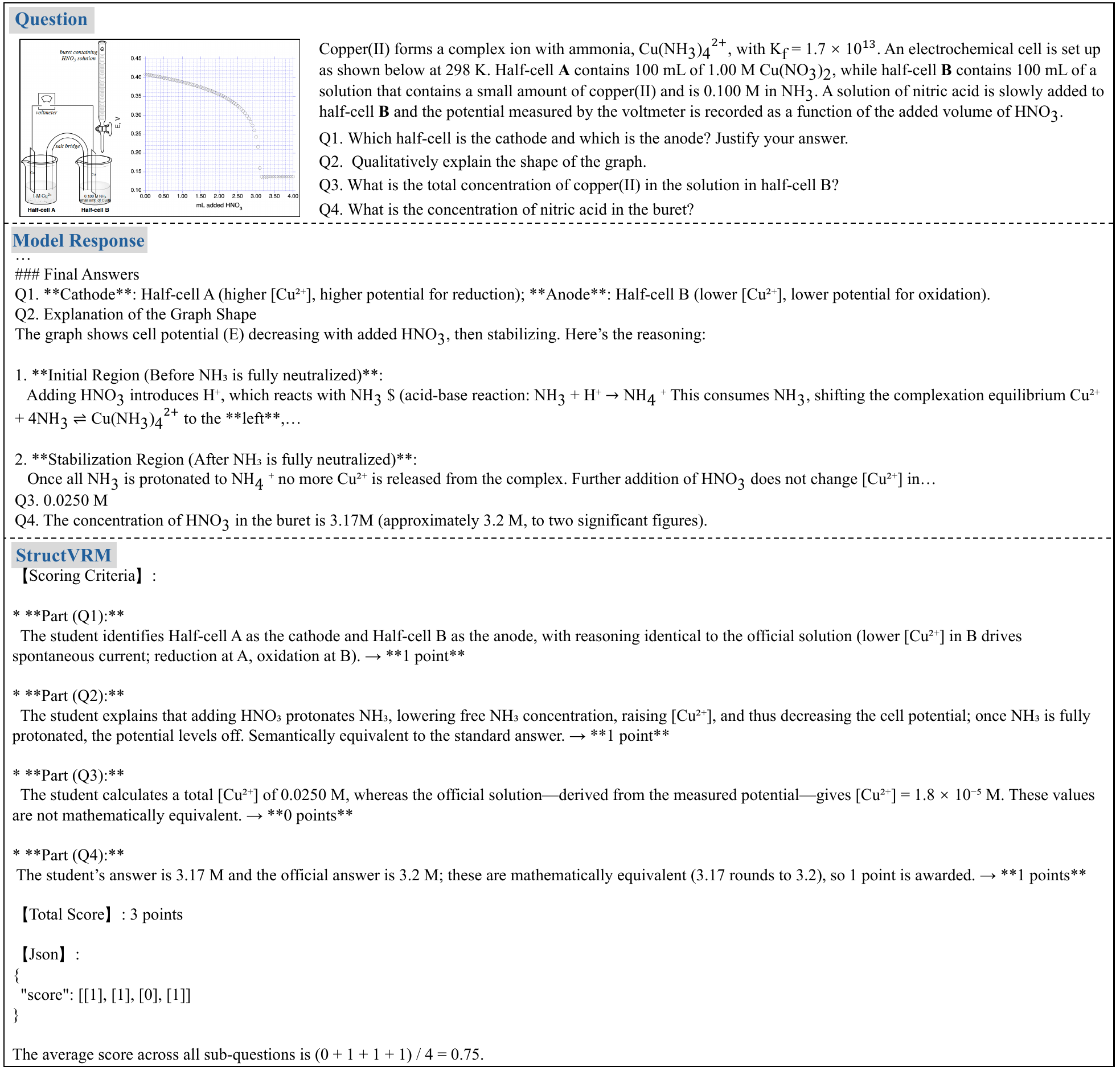}
\caption{StructVRM Verification Example}
\label{fig:verify_case}
\vspace{-2px}
\end{figure}

% where each $s_j$ is a one-dimensional list representing the correctness scores of the $j$-th sub-question(s), with each element being either 0 or 1. Each $s_j$ may correspond to a single sub-question or multiple blanks within a sub-question. The final reward used for RL is then computed as the normalized mean of the entire score vector.
% \begin{equation}
% R_{\text{verifier}}(\hat{y}, y) = \frac{1}{k} \sum_{j=1}^k \text{mean}(s_j)
% \end{equation}
For more complex reasoning tasks—including multi-blank fill-ins and open-ended short answer questions—rule-based approaches are inadequate due to expression diversity and the need for partial credit. To address this, we employ the model-based verifier described in Section~\ref{sec:verifier}, which evaluates the predicted answer through fine-grained, structure-aware judgment at the sub-question level. To ensure accurate reward extraction, the verifier’s output is constrained in a standardized JSON format, as shown in the example in Figure~\ref{fig:verify_case}.

% To ensure accurate reward extraction, the verifier’s output is constrained in a standardized JSON format (see Figure~\ref{fig:verify_case}):

% \begin{figure}[H]
% \centering
% \begin{CJK}{UTF8}{gbsn}
% \begin{center}
% \begin{promptbox}[Verifier Output Format]
% % \begin{lstlisting}[language=json]
% {
%   "score": [[1], [1], [0], [1]]
% }
% % \end{lstlisting}
% \end{promptbox}
% \end{center}
% \end{CJK}
% \caption{Standardized JSON format used by the verifier for structured subquestion scoring.}
% \label{fig:verifier-json}
% \end{figure}

This allows reliable parsing and numerical reward computation:
\begin{equation}
% \texttt{compute\_score}(\texttt{response}) = \min(1, \max(0, \text{mean(flatten(score))}))
\text{Compute Score}(\text{Response}) \;=\;
\text{Clip}\!\bigl (
  \text{Mean}\!\bigl(\text{Flatten}(\text{Score})\bigr),
  \;0,\;1
\bigr)
\end{equation}

Such structured scoring enables reinforcement over previously intractable question formats. For instance, in science problems with multiple blanks spanning factual and procedural knowledge, the verifier issues fine-grained sub-scores reflecting partial correctness across sub-answers. Likewise, in non-verifiable or semi-open questions requiring flexible expressions, our model learns to judge equivalence in meaning and structure, ensuring that genuinely correct answers receive appropriate rewards—even when phrased differently.

This reward mechanism not only stabilizes policy training but also extends applicability to complex, real-world reasoning tasks that defy traditional evaluation methods.
\section{Training Methodology}

% Our SFT and RL training datasets inculde the Seed-1.5-VL general datasets (covering a variety of domains including conversation, creative writing, and general knowledge question answering) and the high-quality, challenging, and verifiable datasets for multimodal reasoning(Section~\ref{sec:Dataset Construction}).

\subsection{Supervised Fine-Tuning}
% Our training pipeline begins with supervised fine-tuning (SFT) based on the Seed1.5-VL model, aiming to instill strong multimodal reasoning capabilities and structured, long-form answer generation. This phase lays the foundation for downstream reward-based optimization in the StructVRM method.
Our training pipeline begins with supervised fine-tuning (SFT), aiming to instill strong multimodal reasoning capabilities and structured, long-form answer generation. This phase lays the foundation for downstream reward-based optimization in the StructVRM method.
% To construct high-quality reasoning trajectories, we adopt a multi-source generation strategy. For each multimodal reasoning instance, we sample diverse responses using five internal models. Each model generates five independent reasoning paths under a unified Chain-of-Thought (CoT) prompt template, designed to elicit explicit step-by-step logic and boxed final answers (Figure~\ref{fig:cot-prompt}). This ensures that outputs are structurally rich and suitable for downstream verification.
To construct high-quality reasoning trajectories, we adopt a multi-source generation strategy. For each multimodal reasoning instance, we sample diverse responses using multiple internal models. Each model generates five independent reasoning paths under a unified Chain-of-Thought (CoT) prompt template, resulting in structurally rich outputs suitable for downstream verification.

% \begin{figure}[t]
% \centering
% \begin{CJK}{UTF8}{gbsn}

% \begin{center}
% \begin{promptbox}[CoT Reasoning Prompt]

% Your answer format needs to be clear and well-organized; your response may include, in order, detailed solution steps, an answer summary, etc. In the answer summary, you must wrap each question’s answer in the format \(\texttt{boxed\{ANSWER HERE\}}\).

% \end{promptbox}
% \end{center}
% \end{CJK}
% \caption{Reasoning prompt used for generating diverse CoT trajectories.}
% \label{fig:cot-prompt}
% \end{figure}

We then perform data quality filtering through a model-based verifier (Section~\ref{sec:verifier}). A lightweight consistency checker evaluates the alignment between each generated answer and its reference label. Based on pass rate statistics, we retain only those responses that fall within a target difficulty band---excluding trivial cases (pass rate > 0.8) and fully failed generations (pass rate = 0). Among the retained candidates, we select the longest correct response as the training target to encourage reasoning depth. 
For harder instances (pass rate = 0), we modify the prompt and repeatedly attempt generation to uncover latent reasoning patterns.
% For harder instances (pass rate = 0), we apply prompt modifications to reattempt generation and unlock latent reasoning patterns.

Prior to CoT generation, all data undergo structural filtering. We remove invalid examples such as those missing diagrams, lacking complete answers, or exhibiting malformed problem descriptions. This rule-based cleaning stage ensures that all training samples contain complete multimodal context and structured outputs.

To further improve alignment between vision and reasoning, we prepend explicit vision prompts to all image-based questions (Figure~\ref{fig:vision-prompt}). These prompts encourage the model to examine the image before reasoning, improving alignment between visual perception and reasoning across complex multimodal tasks.

\begin{figure}[t]
\centering
\begin{CJK}{UTF8}{gbsn}
% \resizebox{\textwidth}{!}{
\begin{center}
\resizebox{\textwidth}{!}{
\begin{promptbox}[Visual Reasoning Prompt]
\ttfamily
Note: 

1.The question contains images; before solving, you need to first interpret the images in conjunction with the textual description, and then describe in detail the key information in the images that is helpful for solving the problem, after which you combine the graphic and textual information to answer. 

2.Your answer format needs to be clear and well-organized; the response may include, in order, detailed solution steps (including the key information from the images that aids in solving the problem), an answer summary, etc. In the answer summary, you must wrap each question’s answer in the format \(\boxed{ANSWER HERE}\).
\end{promptbox}}
\end{center}
\end{CJK}
\caption{Prompt used to encourage vision-aware reasoning in multimodal settings.}
\label{fig:vision-prompt}
\vspace{-4px}
\end{figure}

% We also apply structural augmentation to diversify problem formats. Specifically, we convert a subset of single-answer multiple-choice questions into fill-in-the-blank or binary judgment formats. This discourages shallow pattern matching and forces the model to rely on reasoning over answer semantics, not answer space biases.
The model is fine-tuned on this curated dataset using the AdamW optimizer with hyperparameters $\beta_1=0.9$, $\beta_2=0.95$, and a weight decay of $0.1$; a cosine learning rate schedule decaying from $2\times10^{-5}$ to $2\times10^{-6}$ with a warm-up phase spanning $10\%$ of total steps; a sequence length of $131{,}072$ tokens; and a batch size equivalent to $16\times$ the sequence length. All sequences are truncated to a maximum of $131{,}072$ tokens to preserve long-form reasoning chains. This process results in a model proficient at multimodal reasoning, capable of producing verifier-aligned CoT trajectories that are robust across task structures and modalities.

% The model is fine-tuned for two epochs on this curated dataset using a cosine learning rate schedule (from $2 \times 10^{-5}$ to $2 \times 10^{-6}$). All sequences are truncated to a maximum of 32,000 tokens to preserve long-form reasoning chains. This process results in a model proficient at multimodal reasoning, capable of producing verifier-aligned CoT trajectories that are robust across task structures and modalities.

\subsection{Reinforcement Learning}

Following supervised fine-tuning, we further optimize the model using reinforcement learning (RL) with structured, verifier-aligned reward feedback. The goal of this stage is to refine model behavior toward accurate and format-aware reward supervision across diverse reasoning tasks. Our RL framework focuses on providing structured, verifier-aligned rewards that support partial correctness and semantic-level equivalence in weakly verifiable settings.

To support this, we design format-specific reward mechanisms tailored to the nature of the reasoning output:

\begin{itemize}
    \item \textbf{Multi-choice}: Rule-based verifier. For questions with one or more correct options, predictions are deterministically matched against gold labels. This approach offers reliable signals for structured formats while reducing guessability through multi-answer configurations.
    
    \item \textbf{Structured and open-ended problems}: Model-based verifier. For problems involving multiple subparts or open-ended responses, we adopt the model-based verifier described in Section~\ref{sec:verifier}. It generates a score vector at the sub-question level based on semantic or numerical equivalence, and the final reward is computed as the normalized mean. This supports fine-grained supervision under weakly verifiable conditions and promotes generalization across formats.
\end{itemize}

These strategies follow the reward modeling formulation in Section~\ref{sec:reward-modeling}. The model-based verifier plays a key role in assessing semantically or mathematically equivalent expressions, extending reward coverage beyond rigid matching.

We use distinct KL divergence coefficients for general and verifiable prompts. For general prompts, we apply a small KL coefficient of \(1 \times 10^{-5}\) to prevent potential reward hacking. In contrast, a coefficient of 0 is used for verifiable prompts to allow for greater exploration and flexibility. This approach encourages the model to engage in more exploratory behavior when faced with verifiable tasks, while still maintaining control over general prompts.

For the training process, the context length and maximum output length of RL training are set to 8,192 and 16,384, respectively. In each episode, we sample 4,096 roll-outs, with a mini-batch size of 512 samples, performing 8 gradient steps per episode. The PPO clip range is set to 0.2. Learning rates for the actor and critic are \(6 \times 10^{-7}\) and \(7.5 \times 10^{-7}\), respectively. The number of roll-outs varies depending on the difficulty of the prompt, as harder prompts require more extensive exploration. We sample 4 to 8 times for prompts rewarded by verifiers. This reinforcement learning process, guided by verifier feedback, enables scalable and fine-grained optimization, effectively supporting both rule-based and model-based verification across a variety of reasoning tasks.
\section{Experiments}

\subsection{Experimental Setup}

For our base model, we utilized our internal 20B/200B parameter Mixture-of-Experts (MoE) model. The Seed-StructVRM(our trained model) post-training process comprised an initial SFT stage, followed by a hybrid RL procedure: training prompts are categorized into general and verifiable prompts based on tasks, rewarded with general RM, the rule-based verifier and the model-based verifier(StructVRM).

Our SFT and RL training datasets include diverse general datasets (covering domains such as conversation, creative writing, and general knowledge question answering) as well as high-quality, challenging, and verifiable datasets for multimodal reasoning(Section~\ref{sec:Dataset Construction}).

For evaluation, we evaluate Seed-StructVRM on both standard multimodal reasoning benchmarks and a curated benchmark, STEM-Bench, to comprehensively assess model performance across reasoning formats and difficulty levels. The standard benchmarks target general reasoning accuracy and robustness across modalities. In contrast, STEM-Bench emphasizes subject-specific reasoning, stepwise logic, and structured answer quality under high-fidelity task settings. On standard benchmarks such as VLM2 Bench, EMMA-Mini, ScienceQA, MathVista, CMMMU, MMMU, MMMU-Pro, RealworldQA, and MME, we report accuracy using single-pass inference (pass@1) as the primary metric. For STEM-Bench, we adopt an automated evaluation approach to better assess answer correctness. A large language model (LLM) is used to extract and critique both the model-generated answers and the reference answers. For each sample, the LLM performs three independent rounds of reasoning and scoring, and the final score is obtained by averaging the results. This method ensures more consistent, objective, and scalable evaluation.

\subsection{Main Results on Public Multimodal Benchmarks}

% As shown in Table~\ref{tab:addlabel1}, StructVRM consistently achieves strong performance across a range of multimodal reasoning and visual question answering benchmarks. It ranks first on 6 out of 12 datasets—including VLM2 Bench, Zerobench (sub), ScienceQA, CMMMU, MME Realworld-en, and RealworldQA—and remains highly competitive on the remaining ones. This indicates that the verifier-routed training strategy employed by StructVRM not only enhances structured reasoning but also generalizes effectively across diverse formats and task settings.
As shown in Table~\ref{tab:addlabel1}, Seed-StructVRM consistently achieves strong performance across a range of multimodal reasoning and visual question answering benchmarks. It ranks first on 6 out of 12 datasets, including VLM2 Bench, Zerobench (sub), ScienceQA, CMMMU, MME Realworld-en, and RealworldQA, and remains highly competitive on the remaining ones. This indicates that the verifier-routed training strategy employed by StructVRM not only enhances structured reasoning but also generalizes effectively across diverse formats and task settings.

% In datasets emphasizing complex reasoning and explanation—such as ScienceQA and zerobench—Seed-1.5-VL-StructVRM shows particularly strong results. It achieves 95.1\% accuracy on ScienceQA, outperforming GPT-4o by a large margin, and secures the top score on zerobench (32.6\%), surpassing Claude-Sonnet-4 and Gemini by over 6 points. Even in general visual QA tasks like RealworldQA and MME, where models are typically optimized for broad vision-text understanding, Seed-1.5-VL-StructVRM remains competitive, topping RealworldQA (81.6\%) and performing robustly on both MME-cn and MME-en.
In datasets emphasizing complex reasoning and explanation, such as ScienceQA and Zerobench, Seed-StructVRM shows particularly strong results. It achieves 95.1\% accuracy on ScienceQA, outperforming GPT-4o by a large margin, and secures the top score on Zerobench (32.6\%), surpassing Claude-Sonnet-4 and Gemini by over 6 points. Even in general visual QA tasks like RealworldQA and MME, where models are typically optimized for broad vision-text understanding, Seed-StructVRM remains competitive, topping RealworldQA (81.6\%) and performing robustly on both MME-cn and MME-en.

\begin{table}[h]
  \centering
  \caption{Performance comparison on public multimodal reasoning and visual QA benchmarks.}
  \resizebox{\linewidth}{!}{
    \begin{tabular}{cccc|cccc}
    \toprule
    \multicolumn{1}{c}{Capability} & Benchmark & \cellcolor[rgb]{ .859,  .965,  .984}Seed-StructVRM & Baseline & Claude-Sonnet-4 & o3 & Gemini 2.5 pro & Qwen2.5-VL-72B \\
    \midrule
    \multicolumn{1}{c}{\multirow{9}[18]{*}{\makecell{Multimodal\\Reasoning}}} 
      & VLM2 Bench         & \cellcolor[rgb]{ .859,  .965,  .984}$\textbf{69.8}_{(+5.3)}$ & 64.5 & $45.3^*$ & $63.9^*$ & $69.6^*$ & $62.5^*$ \\
\cmidrule{2-8} 
      & EMMA-Mini          & \cellcolor[rgb]{ .859,  .965,  .984}$57.5_{(+1.5)}$      & 56.0 & $53.8^*$ & $\textbf{64.9}^*$ & $63.5^*$ & $41.0^*$ \\
\cmidrule{2-8} 
      & Zerobench (sub)    & \cellcolor[rgb]{ .859,  .965,  .984}$\textbf{32.6}_{(+1.2)}$ & 31.4 & $24.0^*$ & $25.2^*$ & $26.0^*$ & 13.0 \\
\cmidrule{2-8} 
      & ScienceQA          & \cellcolor[rgb]{ .859,  .965,  .984}$\textbf{95.1}_{(+0.5)}$ & 94.6 & $89.0^*$ & $92.3^*$ & $94.3^*$ & $83.9^*$\\
\cmidrule{2-8} 
      & Mathvision         & \cellcolor[rgb]{ .859,  .965,  .984}$69.5_{(+0.2)}$  & 69.3 & $67.4^*$ & $\textbf{74.0}^*$ & $73.3^*$ & 38.1 \\
\cmidrule{2-8} 
      & CMMMU              & \cellcolor[rgb]{ .859,  .965,  .984}$\textbf{73.2}_{(+0.4)}$ & 72.8 & $66.2^*$ & $71.3^*$ & $70.7^*$ & $58.7^*$ \\
\cmidrule{2-8} 
      & MMMU               & \cellcolor[rgb]{ .859,  .965,  .984}$77.2_{(+0.0)}$    & 77.2 & $75.7^*$ & $\textbf{82.9}^*$ & $81.7^*$ & 70.2 \\
\cmidrule{2-8} 
      & MMMU-pro           & \cellcolor[rgb]{ .859,  .965,  .984}$68.9_{(+1.3)}$      & 67.6 & $65.1^*$ & $\textbf{71.0}^*$ & $68.8^*$ & 51.1 \\
\cmidrule{2-8} 
      & VLMs are Blind   & \cellcolor[rgb]{ .859,  .965,  .984}$\textbf{91.3}_{(+0.5)}$    & 90.8 & $75.5^*$ & $90.0^*$ & $84.3^*$ & 69.0 \\
    \midrule
    \multicolumn{1}{c}{\multirow{3}[6]{*}{General VQA}} 
      & RealworldQA        & \cellcolor[rgb]{ .859,  .965,  .984}$\textbf{81.6}_{(+3.0)}$ & 78.6 & $68.5^*$ & $80.0^*$ & $78.0^*$ & 75.7 \\
\cmidrule{2-8} 
      & MME Realworld-cn   & \cellcolor[rgb]{ .859,  .965,  .984}$67.1_{(+3.0)}$     & 64.1 & $37.9^*$ & $\textbf{71.7}^*$ & $62.0^*$ & $58.0^*$ \\
\cmidrule{2-8} 
      & MME Realworld-en   & \cellcolor[rgb]{ .859,  .965,  .984}$\textbf{67.1}_{(+3.6)}$  & 63.5 & $34.3^*$ & $66.3^*$ & $64.4^*$ & $59.8^*$ \\
    \bottomrule
        \end{tabular}}
  \label{tab:addlabel1}%

\vspace{-1mm}
\small
\raggedright\scriptsize  \ \ *Results self-collected via API in May 2025.\par

\end{table}

This suggests that its structured, reasoning-centric optimization transfers well to open-domain multimodal tasks, and confirms its effectiveness as a general-purpose reasoning model across varied benchmarks.

\subsection{Performance on STEM-Bench}

% We report model performance on STEM-Bench in Table~\ref{tab:stem_bench_results}, which offers a challenging and structured testbed for multimodal scientific reasoning. StructVRM delivers the highest overall score (79.23), outperforming all baselines by a wide margin. Compared to its predecessor Seed1.5-VL (75.51), it shows a consistent gain across all subjects, underscoring the effectiveness of verifier-guided optimization for structured reasoning. Other strong models such as Gemini-2.5 and o3 trail by several points, while generalist models like Claude-Sonnet-4 and Qwen2.5-VL perform significantly lower, especially on solution and fill-in tasks.

We report model performance on STEM-Bench in Table~\ref{tab:stem_bench_results}, which offers a challenging and structured testbed for multimodal scientific reasoning. Seed-StructVRM delivers the highest overall score, outperforming all baselines by a wide margin. Other strong models such as Gemini-2.5-pro and o3 trail by several points, while generalist models like Claude-Sonnet-4 and Qwen2.5-VL perform significantly lower, especially on solution and fill-in tasks.

% From a subject-wise perspective, StructVRM ranks first in all four disciplines—math (93.63), physics (85.78), chemistry (81.22), and biology (88.56). The advantage is most prominent in physics and chemistry, where the model's ability to handle multi-step reasoning and symbolic manipulation is clearly reflected in the fill-in and solution scores. For instance, StructVRM achieves 33.56 on physics solutions, more than five points above Gemini and far beyond GPT-4o (6.22). This demonstrates the model’s superior grasp of procedural reasoning and alignment with human-evaluated correctness. The results collectively validate STEM-Bench as a discriminative benchmark and highlight the effectiveness of StructVRM’s verifier-routed reasoning pipeline in high-difficulty academic domains.
\begin{table}[h]
  \centering
  \caption{Evaluation results on the proposed STEM-Bench across four science subjects.}
  \resizebox{\linewidth}{!}{
    \begin{tabular}{cccc|cccc}
    \toprule
    \multicolumn{2}{c}{Model} & \cellcolor[rgb]{ .859,  .965,  .984}Seed-StructVRM & Baseline & Claude-Sonnet-4 & o3 & Gemini-2.5-pro & Qwen2.5-VL-72B \\

    \midrule
    \multirow{3}[6]{*}{Physics} 
        & MCQ   & \cellcolor[rgb]{ .859,  .965,  .984}$\textbf{48.33}_{(+2.66)}$ & 45.67 & 29.22 & 32.67 & 44.56 & 23.78 \\
\cmidrule{2-8}    & FFQ   & \cellcolor[rgb]{ .859,  .965,  .984}$\textbf{23.78}_{(+5.56)}$ & 18.22 & 15.67 & 18.72 & 23.67 & 5.89 \\
\cmidrule{2-8}    & Total & \cellcolor[rgb]{ .859,  .965,  .984}$\textbf{72.11}_{(+8.22)}$ & 63.89 & 44.89 & 51.39 & 68.22 & 29.67 \\
    \midrule
    \multirow{3}[6]{*}{Chemistry} 
        & MCQ   & \cellcolor[rgb]{ .859,  .965,  .984}$35.89_{(+0.33)}$ & 35.56 & 24.00 & 30.56 & \textbf{36.11} & 23.11 \\
\cmidrule{2-8}    & FFQ & \cellcolor[rgb]{ .859,  .965,  .984}$\textbf{41.22}_{(+2.44)}$ & 38.78 & 28.44 & 27.44 & 38.89 & 19.78 \\
\cmidrule{2-8}    & Total & \cellcolor[rgb]{ .859,  .965,  .984}$\textbf{77.11}_{(+2.78)}$ & 74.33 & 52.44 & 58.00 & 75.00 & 42.89 \\
    \midrule
    \multirow{3}[6]{*}{Biology} 
        & MCQ   & \cellcolor[rgb]{ .859,  .965,  .984}$36.33_{(-0.34)}$ & 36.67 & 22.78 & 29.78 & \textbf{38.00} & 30.56 \\
\cmidrule{2-8}    & FFQ & \cellcolor[rgb]{ .859,  .965,  .984}$\textbf{45.22}_{(+1.33)}$ & 43.89 & 34.44 & 28.89 & 38.72 & 32.78 \\
\cmidrule{2-8}    & Total & \cellcolor[rgb]{ .859,  .965,  .984}$\textbf{81.56}_{(+1.00)} $& 80.56 & 57.22 & 58.67 & 76.72 & 63.33 \\
    \midrule
    \multirow{3}[6]{*}{Math} 
        & MCQ      & \cellcolor[rgb]{ .859,  .965,  .984}$43.04_{(-0.96)}$ & 44.00 & 38.30 & 43.56 & \textbf{44.63} & 29.81 \\
\cmidrule{2-8}    & FFQ & \cellcolor[rgb]{ .859,  .965,  .984}$43.11_{(+3.85)}$ & 39.26 & 37.33 & 34.52 & \textbf{46.81} & 18.07 \\
\cmidrule{2-8}    & Total     & \cellcolor[rgb]{ .859,  .965,  .984}$86.15_{(+2.89)}$ & 83.26 & 75.63 & 78.07 & \textbf{91.44} & 47.89 \\
    \midrule
    \multicolumn{2}{c}{Final Total} & \cellcolor[rgb]{ .859,  .965,  .984}$\textbf{79.23}_{(+3.72)}$ & 75.51 & 57.55 & 61.53 & 77.85 & 45.94 \\
    \bottomrule
    
    \end{tabular}}
  \label{tab:stem_bench_results}

\end{table}
% From a subject-wise perspective, Seed-1.5-VL-StructVRM now leads three of the four STEM-Bench tracks—physics, chemistry, and biology—while securing a strong second place in mathematics. In physics, Seed-1.5-VL-StructVRM scores 72.11, outpacing Gemini 2.5 pro’s 68.22 by nearly four points. In chemistry, the largest advantage appears, with Seed-1.5-VL-StructVRM achieving 77.11 versus 75.00, thanks to its significantly higher free-form-question (FFQ) accuracy of 41.22 compared to 23.78. These FFQs combine several related sub-questions into a single prompt, requiring coherent, multi-part reasoning—an area where Seed-1.5-VL-StructVRM excels. In biology, Seed-1.5-VL-StructVRM scores 81.56 against the next-best model’s 80.56, showing that its sub-question decomposition strategy generalizes beyond strictly quantitative domains. Although Seed-1.5-VL-StructVRM places second in mathematics with 86.15 versus Gemini 2.5 pro’s 91.44, it still comfortably outperforms all other baselines. Averaging 79.23 across STEM-Bench, Seed-1.5-VL-StructVRM sets a new benchmark high, underscoring both the discriminative power of STEM-Bench and the strength of Seed-1.5-VL-StructVRM’s decomposition-based pipeline in tackling complex academic tasks.

From a subject-wise perspective, Seed-StructVRM now leads three of the four STEM-Bench tracks: physics, chemistry, and biology, while maintaining a strong second-place position in mathematics. In physics, it scores 72.11, outperforming Gemini-2.5-Pro’s 68.22 by nearly four points. The largest margin appears in chemistry, where the model achieves 77.11 compared to 75.00, driven by a significantly higher free-form question (FFQ) accuracy of 41.22 versus 23.78. These FFQs combine several related sub-questions into a single prompt and demand coherent, multi-question reasoning—an area in which the model excels. In biology, it reaches 81.56, slightly surpassing the next-best score of 80.56, demonstrating that its sub-question decomposition strategy generalizes well beyond quantitative domains. Although it places second in mathematics with a score of 86.15, compared to Gemini-2.5-Pro’s 91.44, it still outperforms all other baselines by a notable margin. With an average score of 79.23 on STEM-Bench, the model sets a new benchmark high, underscoring the benchmark’s discriminative strength and validating the impact of Structured Rewards in solving complex academic challenges.

% With an average score of 79.23 across STEM-Bench, the model sets a new benchmark high, highlighting both the discriminative power of the benchmark and the effectiveness of its decomposition-based pipeline in addressing complex academic challenges.

% \subsection{Ablation Study}

% \begin{table}[t]
%   \centering
%   \caption{Ablation results on STEM-Bench.}
%   \resizebox{\linewidth}{!}{
%     \begin{tabular}{c|ccc|ccc|ccc|ccc|c}
%     \toprule
%     \multirow{2}[2]{*}{Model} & \multicolumn{3}{c|}{Math} & \multicolumn{3}{c|}{Physics} & \multicolumn{3}{c|}{Chemistry} & \multicolumn{3}{c|}{Biology} & \multirow{2}[2]{*}{Final Total} \\
%           & MCQ   & Solutions & Total & MCQ   & Solutions & Total & MCQ   & Fill-in & Total & MCQ   & Fill-in & Total &  \\
%     \midrule
%     Seed-1.5-VL-StructVRM & \textbf{68.89} & \textbf{71.56} & \textbf{93.63} & \textbf{52.22} & \textbf{33.56} & \textbf{85.78} & \textbf{40.33} & \textbf{40.89} & \textbf{81.22} & \textbf{39.67} & \textbf{48.89} & \textbf{88.56} & \textbf{87.30} \\
%     w/o Verifier & 68.00    & 72.22 & 93.48 & 48.44 & 32.89 & 81.33 & 40.00    & 40.00    & 80.00    & 37.78 & 47.33 & 85.11 & 84.98 \\
%     w/o RL & 67.61 & 67.11 & 89.81 & 47.78 & 33.00    & 80.78 & 38.67 & 40.89 & 79.56 & 38.33 & 46.89 & 85.22 & 83.84 \\
%     \bottomrule
%     \end{tabular}}
%   \label{tab:ablation-results}
% \end{table}%

\subsection{Ablation Study}

We conduct ablation experiments to quantify the contribution of key components in the Seed-StructVRM training pipeline: the StructVRM and reinforcement learning. Table~\ref{tab:ablation-results} summarizes performance under three configurations: full model, removal of the StructVRM, and removal of reinforcement learning.

\begin{table}[h]
  \centering
  \caption{Ablation results on STEM-Bench.}
  \resizebox{\linewidth}{!}{
    \begin{tabular}{c|ccc|ccc|ccc|ccc|c}
    \toprule
    \multirow{2}[2]{*}{Model} & \multicolumn{3}{c|}{Math} & \multicolumn{3}{c|}{Physics} & \multicolumn{3}{c|}{Chemistry} & \multicolumn{3}{c|}{Biology} & \multirow{2}[2]{*}{Final Total} \\
          & MCQ & FFQ & Total & MCQ & FFQ & Total & MCQ & FFQ & Total & MCQ & FFQ & Total &  \\ 
    \midrule
    Seed-StructVRM   & 43.04 & \textbf{43.11} & \textbf{86.15} & \textbf{48.33} & \textbf{23.78} & \textbf{72.11} & \textbf{35.89} & \textbf{41.22} & \textbf{77.11} & 36.33 & \textbf{45.22} & \textbf{81.56} & \textbf{79.23} \\
    w/o StructVRM   & 43.26 & 41.26 & 84.52 & 47.78 & 22.44 & 70.22 & 35.44 & 39.11 & 74.56 & 35.22 & 42.11 & 77.33 & 76.66 \\
    w/o RL         & \textbf{44.81} & 38.00 & 82.81 & 47.33 & 21.61 & 68.94 & 34.89 & 37.22 & 72.11 & \textbf{36.44} & 41.56 & 78.00 & 75.47 \\
    \bottomrule
    \end{tabular}}
  \label{tab:ablation-results}
\end{table}

The full Seed-StructVRM configuration achieves the best overall score (79.23), confirming the benefit of combining structured reward feedback with policy optimization. Removing the StructVRM results in a moderate drop to 76.66, whereas removing RL leads to a larger decline to 75.47. These findings suggest that while the StructVRM helps fine-tune scoring alignment, reinforcement learning plays a more pivotal role in enhancing multi-question reasoning robustness.

Subject-level trends reinforce this: without RL, math total drops from 86.15 to 82.81 and chemistry total from 77.11 to 72.11, underscoring RL’s impact on structured derivation and long-form problem solving. The absence of the StructVRM causes lighter but consistent reductions, especially in physics (from 72.11 to 70.22) and biology (from 81.56 to 77.33), suggesting it enhances reliability in evaluating ambiguous or partially correct answers.

These results suggest that the two components improve different aspects of performance: reinforcement learning notably enhances multi-question reasoning accuracy, while the StructVRM contributes to more reliable scoring, particularly in open-form questions where partial correctness must be evaluated.

% \subsection{Qualitative Case Study}
% \begin{figure}[!t]
%     \centering
%     \includegraphics[width=\textwidth]{images/case11.pdf}
%     \caption{Multi-question reasoning and single-question reasoning.}
%     \label{fig:demo1}
% \end{figure}

% \begin{figure}[t]
%     \centering
%     \includegraphics[width=\textwidth]{images/demo2.pdf}
%     \caption{Spatial quantity counting and multi-question physics reasoning.}
%     \label{fig:demo2}
% \end{figure}

\subsection{Qualitative Case Study}
\begin{figure}[!ht]
    \centering
    \includegraphics[width=\textwidth]{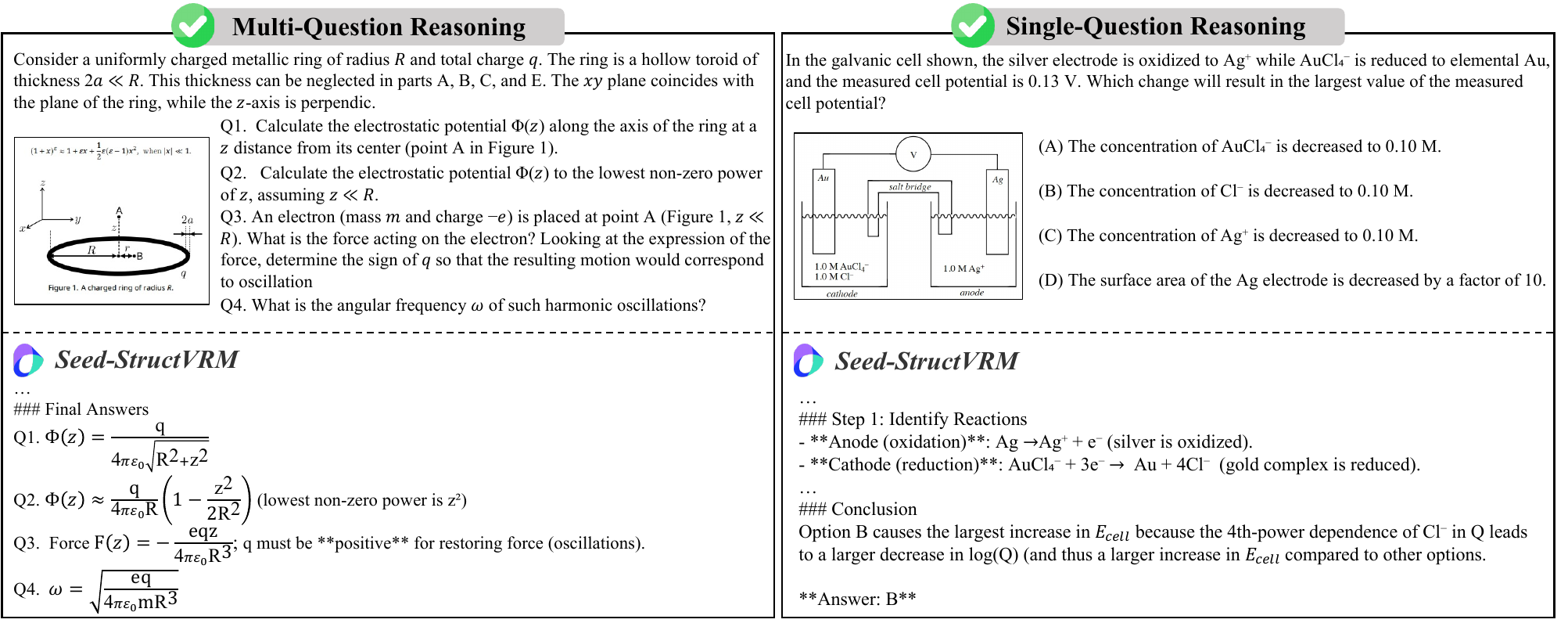}
    \caption{Multi-question reasoning and single-question reasoning.}
    \label{fig:demo1}
    \vspace{-8px}
\end{figure}

Figure~\ref{fig:demo1} presents two examples that further illustrate the scope and robustness of Seed-StructVRM multimodal reasoning capabilities, especially in scientific problem-solving contexts that demand symbolic manipulation, mathematical modeling, and domain-specific causal inference.

In the left-side example of Figure~\ref{fig:demo1}, the model addresses a multi-part theoretical physics problem involving electrostatics and harmonic motion of a charged particle. It begins by computing the electrostatic potential along the ring axis, correctly applying symmetry principles and integral reasoning. Subsequent sub-questions require Taylor expansion for near-field approximations, derivation of electric force via gradient operations, and identification of oscillatory behavior through Newtonian dynamics. Seed-StructVRM navigates this symbolic chain question-by-question, maintaining physical consistency and accurate formula derivation across individual subquestions, and showcasing compositional reasoning grounded in visual schematics and mathematical formalism.

In the right-side case (Figure~\ref{fig:demo1}, right), the model analyzes a single-choice chemistry competition question involving redox reactions in a galvanic cell. Despite the absence of multi-question guidance, Seed-StructVRM autonomously identifies the relevant half-reactions, constructs the overall cell equation, and applies the Nernst equation to assess the impact of different concentration changes. 
\begin{figure}[!ht]
    \centering
    \includegraphics[width=\textwidth]{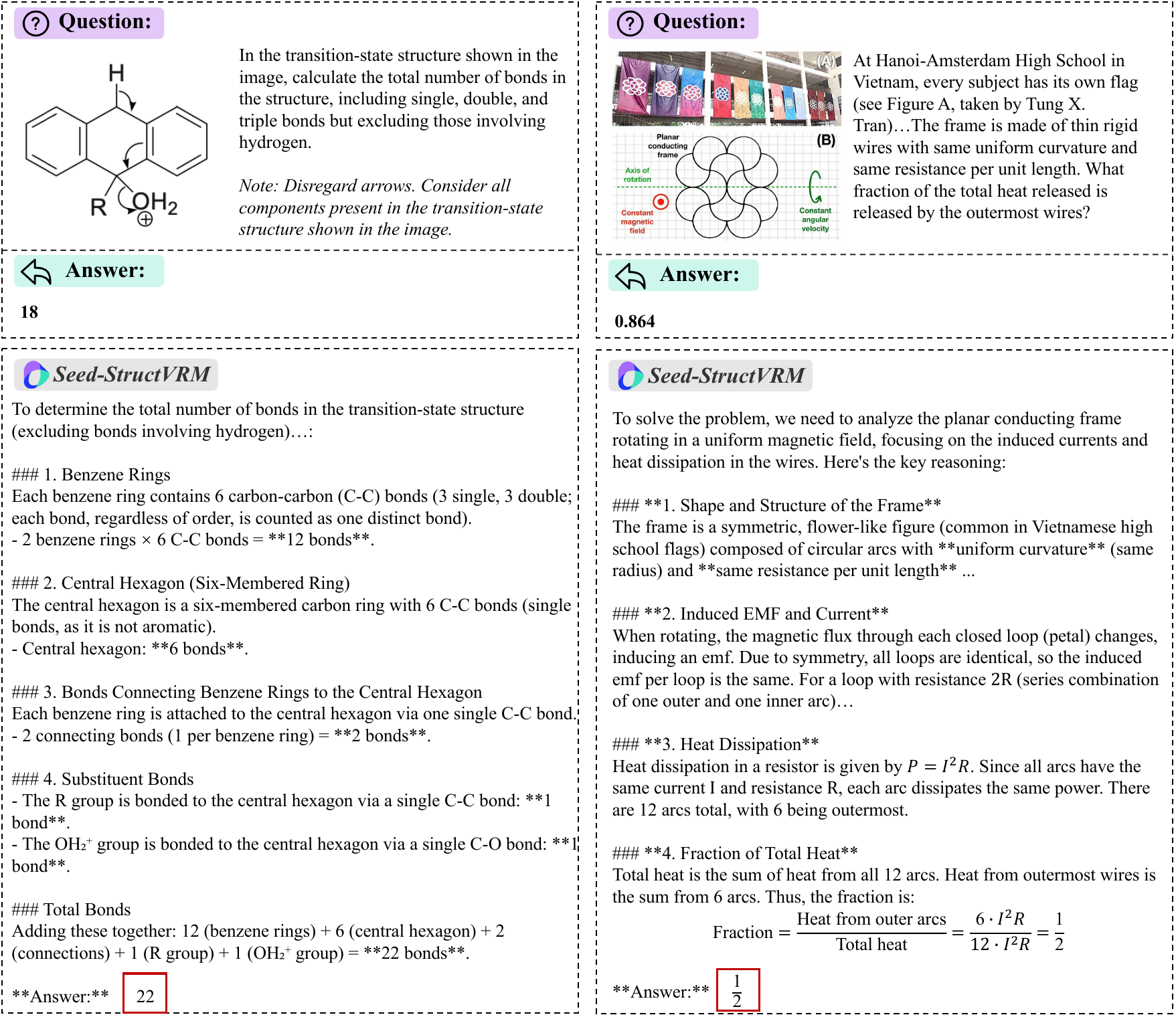}
    \caption{Error case.}
    \label{fig:error2}
    \vspace{-8px}
\end{figure}
% It correctly computes the magnitude of each option's effect on cell potential, recognizing the exponential sensitivity of the reaction quotient to $[\text{Cl}^-]^4$. This reflects the model's capacity to handle both multi-part scientific questions and standalone, domain-specific problems, demonstrating its versatility across decomposable and single-question reasoning tasks alike.
It correctly computes the magnitude of each option's effect on cell potential, recognizing the exponential sensitivity of the reaction quotient. This reflects the model's capacity to handle both multi-part scientific questions and standalone, domain-specific problems, demonstrating its versatility across decomposable and single-question reasoning tasks alike.

Together, these examples reaffirm that Seed-StructVRM performs precise and interpretable multimodal reasoning, whether on multi-question tasks or single, analytically rich problems. This highlights the generalizability of its verifier-guided alignment and structured reasoning supervision in real-world problem solving.

\subsection{Error Analysis}

To better understand the limitations of Seed-StructVRM in complex reasoning scenarios, we examine representative failure cases sampled from STEM-Bench, focusing on high-difficulty questions across disciplines.

% Figure~\ref{fig:error1} shows a case of visual-semantic misalignment. The model fails to recognize that the distinct red/green/blue fluorescence channels indicate a fluorescent staining procedure. Instead, it mistakenly interprets the visualization as conventional visible light staining. This error suggests insufficient grounding between domain-specific visual cues and scientific knowledge, especially when the task requires cross-modal alignment of image features and biochemical techniques.

In Figure~\ref{fig:error2} (left), the model’s reasoning breaks down during bond-counting in an organic molecule. While the correct approach involves systematically identifying all carbon–carbon and carbon–oxygen bonds across cyclic and side-chain structures, Seed-StructVRM undercounts by failing to fully decompose the structural diagram. This points to a lack of precise symbolic parsing in visual-to-chemical structure translation, especially for tasks involving nested substructures.

Figure~\ref{fig:error2} (right) demonstrates a physics reasoning failure. Although the model correctly observes that the outer and inner loops of a resistor wire have equal length, it erroneously assumes the segments are symmetric and skips detailed thermal calculations. This shortcut undermines its ability to handle energy distribution tasks that require exact spatial differentiation and heat transfer modeling.

% In Figure~\ref{fig:error4}, StructVRM struggles with spatial planning in a grid-based vehicle puzzle. The model gives an answer of “4” vehicles needing to move, but detailed step-by-step reasoning shows that only two moves are required. The mistake arises from inconsistent spatial indexing—confusing the top and bottom rows—which leads to unnecessary motion estimates. This highlights the need for more robust topological memory and geometric state tracking.

% Finally, Figure~\ref{fig:error5} reveals a failure in basic quantitative comparison. Despite the number of green particles and solvent volume being identical in both beakers, the model incorrectly claims that solution A is more concentrated. This suggests either visual miscounting or confusion in applying the concentration definition, underlining issues in combining visual enumeration with elementary quantitative reasoning.

These cases expose representative limitations in Seed-StructVRM’s current reasoning process: challenges in grounding domain-specific visual semantics, partial failures in structural parsing of diagrams, over-reliance on shallow heuristics, and inconsistencies in spatial or numerical reasoning. Future work could explore more fine-grained visual-language alignment, improved diagram interpretation modules, and task-specific consistency constraints to address these gaps.
\section{Conclusion}

We present StructVRM, a training and verification method tailored for challenging multimodal reasoning tasks. It enhances reasoning reliability through a verifiable reward model and improves learning stability via structured data augmentation. StructVRM demonstrates strong performance across multiple reasoning benchmarks and offers a scalable foundation for fine-grained supervision in complex, real-world settings.

\newpage
\section{Contributions and Acknowledgments}
\label{sec:contributions}

\subsection*{Core Contributors}

Xiangxiang Zhang, Jingxuan Wei, Donghong Zhong, Qi Chen, Caijun Jia, Cheng Tan, Jinming Gu, Xiaobo Qin

\subsection*{Contributors}

Zhiping Liu,  Liang Hu, Tong Sun, Yuchen Wu, Zewei Sun, Chenwei Lou, Hua Zheng, Tianyang Zhan, Changbao Wang, Shuangzhi Wu,  Zefa Lin,  Chang Guo, Sihang Yuan, Riwei Chen, Shixiong Zhao, Yingping Zhang, Gaowei Wu,  Bihui Yu, Jiahui Wu, Zhehui Zhao, Qianqian Liu, Ruofeng Tang, Xingyue Huang, Bing Zhao, Mengyang Zhang, Youqiang Zhou

\subsection*{Correspondence}

zhangxiangxiang.zxx@bytedance.com, chengtan9907@gmail.com, weijingxuan20@mails.ucas.edu.cn

\subsection*{Affiliation}

ByteDance Seed China \\
Shenyang Institute of Computing Technology, Chinese Academy of Sciences

\clearpage

\bibliographystyle{unsrt}
\bibliography{main}

% \clearpage
% \beginappendix
% \input{sections/appendix}

\end{document}